\documentclass[sn-mathphys-num]{sn-jnl}

\usepackage{graphicx}%
\usepackage{multirow}%
\usepackage{amsmath,amssymb,amsfonts}%
\usepackage{amsthm}%
\usepackage{mathrsfs}%
\usepackage[title]{appendix}%
\usepackage{xcolor}%
\usepackage{textcomp}%
\usepackage{manyfoot}%
\usepackage{booktabs}%
\usepackage{algorithm}%
\usepackage{algorithmicx}%
\usepackage{algpseudocode}%
\usepackage{listings}%

\theoremstyle{thmstyleone}%
\newtheorem{theorem}{Theorem}
\theoremstyle{thmstyletwo}%

\theoremstyle{thmstylethree}%

\begin{document}

\title{Hacking a surrogate model approach to XAI}


\author*[1]{\fnm{Alexander} \sur{Wilhelm}}\email{alexander.wilhelm@dfki.uni-kl.de}

\author[2]{\fnm{Prof. Katharina A.} \sur{Zweig}}\email{zweig@cs.uni-kl.de}
\affil*[1]{\orgname{German Research Center for Artificial
Intelligence (DFKI)}, \orgaddress{\street{Trippstadter Str. 122}, \city{Kaiserslautern}, \postcode{67663}, \country{Germany}}}

\affil*[2]{\orgdiv{Department of Computer Science}, \orgname{RPTU Kaiserslautern-Landau}, \orgaddress{\street{Gottlieb-Daimler-Straße 47}, \city{Kaiserslautern}, \postcode{67663}, \country{Germany}}}


\abstract{In recent years, the number of new applications for highly complex AI systems has risen significantly. Algorithmic decision-making systems (ADMs) are one of such applications, where an AI system replaces the decision-making process of a human expert. As one approach to ensure fairness and transparency of such systems, explainable AI (XAI) has become more important. One variant to achieve explainability are surrogate models, i.e., the idea to train a new simpler machine learning model based on the input-output-relationship of a black box model. The simpler machine learning model could, for example, be a decision tree, which is thought to be intuitively understandable by humans. However, there is not much insight into how well the surrogate model approximates the black box.

Our main assumption is that a good surrogate model approach should be able to bring such a discriminating behavior to the attention of humans; prior to our research we assumed that a surrogate decision tree would identify such a pattern on one of its first levels. However, in this article we show that even if the discriminated subgroup - while otherwise being the same in all categories - does not get a single positive decision from the black box ADM system, the corresponding question of group membership can be pushed down onto a level as low as wanted by the operator of the system.

We then generalize this finding to pinpoint the exact level of the tree on which the discriminating question is asked and show that in a more realistic scenario, where discrimination only occurs to some fraction of the disadvantaged group, it is even more feasible to hide such discrimination.

Our approach can be generalized easily to other surrogate models. 
}




\maketitle

\section{Introduction}
In recent years, the use of \textit{artificial intelligence (AI)} for various tasks has flourished. AI is especially increasingly used to support decision-making or making  decisions in complex situations - these systems are called \textit{algorithmic decision-making systems} or \textbf{ADM systems}. While AI can help to make decision-making faster, the logic involved behind the decisions of these systems is often too complex to be reasonably understandable. However, if such systems are used to make decisions that directly affect human lives, an explanation of how these decisions are made is indispensable and might even be required by law: In Europe, persons subject to automated decision-making have a right to be informed about the decision logic involved~\cite{gdpr}.

For this, different so-called \textit{explainability} approaches have been explored to explain their decision-making process. One such approach is the \textit{surrogate model} approach for classifiers, which assign a class label to input data. A surrogate model is then a simple statistical model, which is trained with data points that have been labeled by the ADM system. The goal of such surrogate models is to gain insight into the black box model by studying the logic involved of the surrogate model~\cite[p.~575]{masch}. Linear regressions and decision trees are considered easy to understand surrogate models because of their intuitive, easy to read representation as a simple formula and a tree, respectively. They are thus often called \textit{white box models}~\cite{bmwi2021},~\cite{rokach2005},~\cite{kotsiantis2013}.

In most cases, linear regression will not be suitable as a surrogate model as most decision rules are not linearly dependent on input features. Thus, it is likely, that decision trees might be considered as a better choice by regulators. However, to avoid information overload, a regulator might come up with the idea that any accompanying decision tree needs to be limited in height or number of nodes to ensure that only the most important decisions are contained.\\

In this paper, we prove mathematically for a simple setting that it is possible to hide obvious discriminatory decisions in the black box model without detecting them in the surrogate decision tree model. We showcase this with a scenario in which a malicious operator uses an ADM system to make decisions about the creditworthiness of their clients. This operator wants to discriminate against certain clients, in the knowledge that they have to produce such an explaining decision tree. The findings can also help regulators to improve their testing processes.\\

We start with necessary definitions and concepts in Sec.~\ref{definition} before we describe the general scenario in Sec.~\ref{DMcredt}. We prove three theorems about this scenario in Sec.~\ref{order}, whose implications are discussed in Sec.~\ref{Implications}, together with future research questions. The paper is concluded in Sec.~\ref{conclusion}.

\section{\label{definition}Definition}
In the following chapters, we introduce basic concepts and terminology for classification and decision trees.

\subsection{\label{def_classification} Classification of a Data Set}
\textit{Classification} is a process in which data points are divided into groups, according to their \textit{attributes} or \textit{features}~\cite[p. 8]{harrington2012}.  An example is the classification of birds into their respective species based on features (or attributes) such as size, their wingspan, the shape of their feet, and their color~\cite[p. 7f.]{harrington2012}. The first two attributes can take on real numbers, the second two are assigned sets of describing words. There is a fifth attribute, the species, that is assigned the set of all known species names.  The decision of which species a bird belongs to is a so-called \textit{label}. 

In general, a data set is a set $S=\{(\vec{x}_1,y_1),(\vec{x}_2,y_2), ...,(\vec{x}_n,y_n)\}$ of data points, consisting of attributes $\vec{x}_j$ and labels $y_j$. The value of the $i$-th attribute can be found at $x_j[i]$. Each attribute is assigned a set $A_i$ of possible values; the set of possible values of the label is called $Y$. In the following we will assume that the label is binary, i.e., $|Y|=2$.

\textit{Artificial intelligence (AI)} or \textit{machine learning (ML)} is defined as the improvement of the performance of an algorithm through experience~\cite[p. 2]{mitchell1997}. For the specific task of classification, such algorithms are called \textit{algorithmic decision-making systems (ADM systems)}. ADM systems use data points and their attributes as their input and assign a label as the output of the system~\cite[p.~430]{masch}. They consist of a statistical model, which is trained using labeled input data, the so-called \textit{training data set}. The systems are also called \textit{black box models} because the underlying statistical model of ADMs and, therefore, the way they assign a label to a data point is often not comprehensible for a human being~\cite[p.~575]{masch}.

\begin{figure}[h!]
    \centering
    \includegraphics[width=0.6\textwidth]{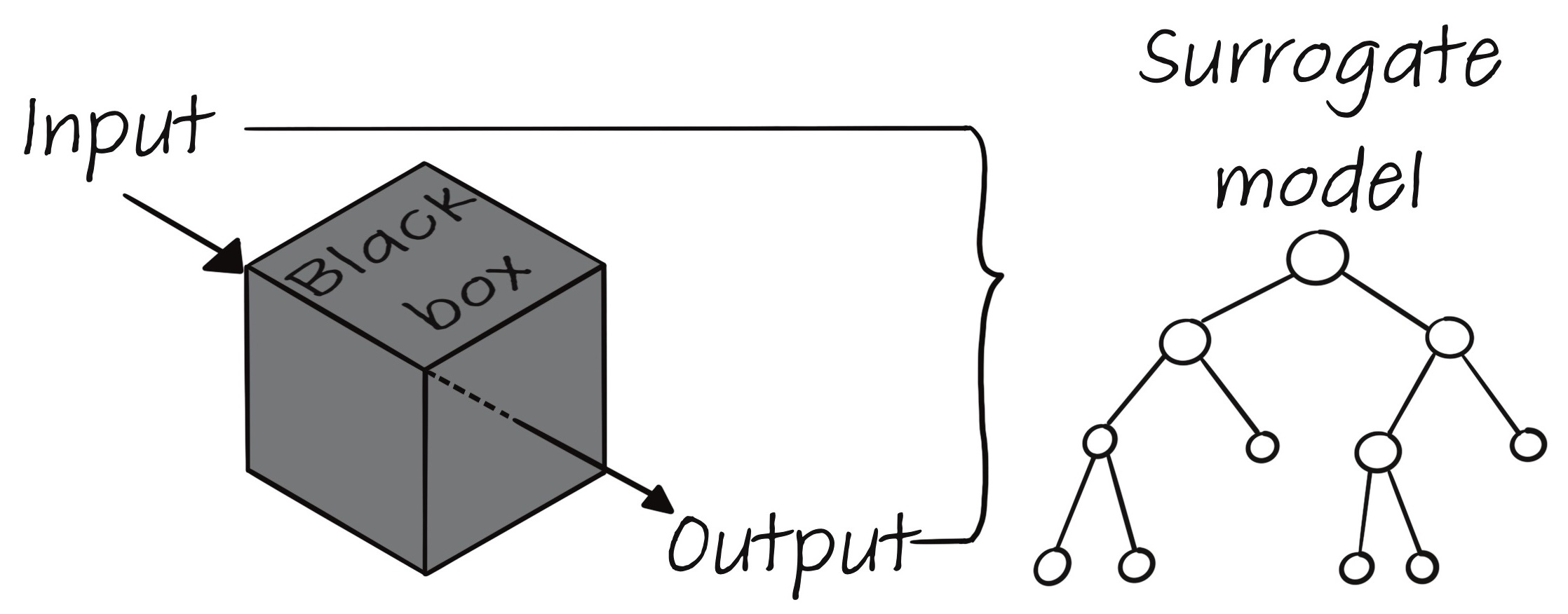}
    \caption{A  surrogate  model  is  trained  with  an  input  data  set  and  its  labels,  which  are assigned by the black box model. It is assumed that the surrogate model thereby approximates  the  logic  involved  of  the  black  box  model.  The  surrogate  model  is then used to try to explain the inner logic of the black box model.}
    \label{surrogate}
\end{figure}

In order to be able to make statements about the inner logic of such models, \textit{surrogate models} are used. A surrogate model is also a statistical model, designed to apply labels through machine learning. In contrast to the black box model, a surrogate uses a statistical model whose logic involved is thought to be so intuitive that it can instantly be understood by humans, a so-called \textit{white box model}. The surrogate model is trained with data points that have been labeled by the black box model, see Fig.~\ref{surrogate}. The goal of such surrogate models is to gain insight of the black box model by studying the inner logic of the surrogate model~\cite[p.~575]{masch}.

An example for such a surrogate model is a decision tree, which is described in the following.

\subsection{Decision Trees}\label{treeex}
\textit{Decision Trees} are directed, ordered trees that represent decision rules~\cite{rokach2005}. These decision rules are functions $f_T:X\rightarrow \{1,2\}$ that assign a  label to the attributes of a data point~\cite[p. 52]{mitchell1997}. In other words, the classification of data points is done via decision rules in a decision tree. A decision tree shows the decision rules graphically, making them more comprehensible~\cite[p.12]{rokach2014}. 

Decision trees consist of a \textit{root node} with only outgoing edges, \textit{inner nodes} and \textit{leaf nodes.} Both the root and inner nodes represent individual decisions that are made based on an attribute. The leaf nodes, on the other hand, contain the classification into various groups.

The decision rules, which are represented in the inner nodes, can be phrased in a sentence with `If... then...'~\cite[p. 52]{mitchell1997},~\cite[p. 45ff.]{quinlan1993}. Such decisions can be binary if the corresponding attribute has only two values. An example of a binary decision rule would be: `If a person has a salary of more than \$$x$ per month, then they should be classified as creditworthy.' The node representing this decision rule would ask for the attribute `salary' by the rule `Has salary over \$$x$ per month', as shown in Fig.~\ref{treebspl}. The subtrees that start from this node then contain further decision rules that are linked to the first rule by an `and'. Traversing from the root node, for each of the leaf nodes, decision rules of the form `If the data point has attribute A with value b and attribute C with value d and..., then it is to be classified as group e.' can be formulated. Thus, any path from the root to a leaf constitutes a decision rule in conjunctive normal form (CNF). The tree represents a set of these rules in a compact manner and displays them with a graphical representation.\\

While it is possible to extract each decision rule from a decision tree, their maximum number is equal to the number of leaf nodes. For a complete binary decision tree of height $h$, the number of decision rules is therefore $2^h$. As the number of decision rules grows, it becomes more and more complex to retain an overview of the decision-making of the tree.

\subsection{Learning Decision Trees From Data}
There are various approaches to build a decision tree from labeled data where one of the attributes is the classification goal; in the scenario showcased here, it would be the creditworthiness. The idea is that the methods identify the most important decision rules that best approximate the labels of the data points in the data set.

However, construction of an optimal decision tree is NP-complete~\cite{hyafil76}, and construction of a minimal binary tree or finding a minimal tree equal to an existing tree are NP-Hard~\cite{zantema00},~\cite{hancock96}. Thus, in the following, the CART algorithm~\cite{rokach2005},~\cite{breiman1984}, a greedy top-down approach, will be examined, which does not always produce an optimal tree~\cite{zantema00}. In this approach, the decision used in the first level of the tree is determined first, followed by the decisions in subsequently lower levels.

In that sense, each node in the decision tree is assigned to a subset of data points: At the root node $r$, it is $S$. Let $S(v)$ denote the assigned data set of node $v$, i.e., $S(r)=S$. For any node $v$ in the tree,  let $q(v)$ denote the attribute in its decision question with values $v_1, v_2, ..., v_n$.  In each of the nodes $v$ linked to some node $w$ and associated with the value $i$ of $q(w)$'s attribute, the assigned data set contains only those elements of $S(w)$ takes on value $i$.  In the creditworthiness scenario, $S$ itself will contain many data points from both groups. The idea is to find the attribute and the corresponding decision question that best splits the data set into ``purer'' groups, i.e., ones that mainly contain one group or the other.\\ 

Now, for each new node, all possible attributes are examined on whether splitting the data set according to their possible values will result in subgroups that are ``purer'' than the data set in full---this is the greedy aspect of the method. There are various possible measures to quantify the ``pureness'' of a data set. In the following, the so-called \textit{Gini index} is used as \textit{splitting criterion}, i.e. as a criterion according to which the ``best'' decision rule is chosen in each node. We chose this method, as it can at least be shown that, locally, the choice it makes, is the best possible\footnote{``Growing a tree by using the Gini splitting rule continually minimizes the resubstituion estimate $R(T)$ for the MSE.''~\cite[p.124]{breiman1984}.}~\cite{breiman1984}.

\subsection{\label{def_gini}Gini Index}
The \textit{Gini index} is a statistical measure that quantifies the inequality of the distribution of a label\footnote{Rokach et al. use a different notation for the Gini index, that has been simplified for this paper~\cite{rokach2005}.}~\cite{rokach2005}. For a label $y$ and a data set $S$, it is defined as:
\begin{equation}
    \text{Gini }(y,S)=1-\sum_{i\in \text{dom}(y)}{p_{i}}^2
\end{equation}
where $p_{i}$ is the fraction of elements in data set $S$ that have a value of $i$ for label $y$, i.e. $p_i:=\frac{\left|\{d_j\in S | y_j = i\}\right|}{| S |}$. This will also be denoted as \textbf{$_S^{y}p_i$}. For creditworthiness, label $y$ has two outcomes: `creditworthy' and `not creditworthy'.

The Gini index quantifies the heterogeneity of $S(v)$, the data set of any given node $v$ inside a decision tree. To determine which split should be chosen for a node, the \textit{weighted Gini impurity after a split}, abbreviated as \textit{Gini impurity} in the following, is used\footnote{In literature, Gini Gain is often used to determine the most optimal split. This measure takes the Gini index of a node and subtracts the Gini impurity of a split from it. Because the Gini index is the same for all possible splits inside a node, both Gini Gain and Gini impurity can be easily exchanged in the calculations shown in this paper.}. It measures the quality of a split based on attribute $a$:
\begin{equation}\begin{split}
    \text{G}(a,S) = \sum_{i\in \text{dom}(a)}p_{i}\cdot \text{Gini }(y,S')
\end{split}\end{equation}
where $S'$ is the set of elements that are in $S$ and have value $i$ for attribute $a$.

The attribute that achieves the lowest possible Gini impurity is selected for a split. This means, a split is performed based on the decision which results in the most homogeneous sub-data sets.

The Gini index and Gini impurity can assume values from $0$ to $0.5$. A Gini index of $0.5$ arises if a set contains equally sized subsets of data points for each value of $y$. I. e., for a binary label, the index is maximal if both values make up half of the data points each. On the other hand, a value of $0$ means that all data points inside a data set are assigned the same label.

Therefore, a high value of the Gini impurity means that all data sets after a split contain a high heterogeneity, weighted according to their size. A low value means that those data sets or the majority of them is relatively homogeneous.\\

\begin{algorithm}
\label{alg}
 \begin{algorithmic}[1]
 \Require Data partition S\\
 a list of attributes
 \Ensure A decision tree
 \State N = createNode()
 \If{stoppingCriterion(S, attributes) is true} \State {N.label = classify(S) //gives the majority class in S} \EndIf
 \State splittingCriterion = $\text{argmin}_{\text{a}\in \text{attributes}}$G(S,a)
 \State N.label = splittingCriterion
 \ForAll{values $i$ of splittingCriterion} \State{attach node returned by GenerateDecisionTree($\text{S}_\text{i}$,attributes-splittingCriterion)} \EndFor\\
 \Return N
 \end{algorithmic}
 \caption{An algorithm to built a decision tree with the Gini index based on the CART algorithm~\cite{breiman1984}.}
\end{algorithm}
Algorithm~\ref{alg} describes how a decision tree is built using the Gini index. It uses the sample data set $D$ and a list of attributes as inputs. A helper function $stoppingCriterion(D, attributes)$ checks whether a stopping criterion, a rule which indicates if a node is to be split further, in a list of stopping criteria is fulfilled and returns true if at least one is fulfilled.\\

Based on this algorithm, we will now explore whether it is possible for a malicious actor to hide discrimination in the automatically deduced decision tree used as a surrogate model. 

\section{\label{DMcredt}Decision-Making on Creditworthiness}

The scenario considered in the following is that an operator, e. g. the director of a bank, decides on credits in a discriminatory and intransparent fashion, i.e., by a black box algorithm. Note that, to avoid any discussions on discrimination triggered by real-world examples, we use fictional groups of creatures (elves and ogres) with coins as their currency---we will come back to the political implications of our findings in Sec.~\ref{Implications}. In this paper, we assume that ogres and elves have an equal salary distribution.

\begin{figure}[h!]
    \centering
    \includegraphics[width=0.8\textwidth]{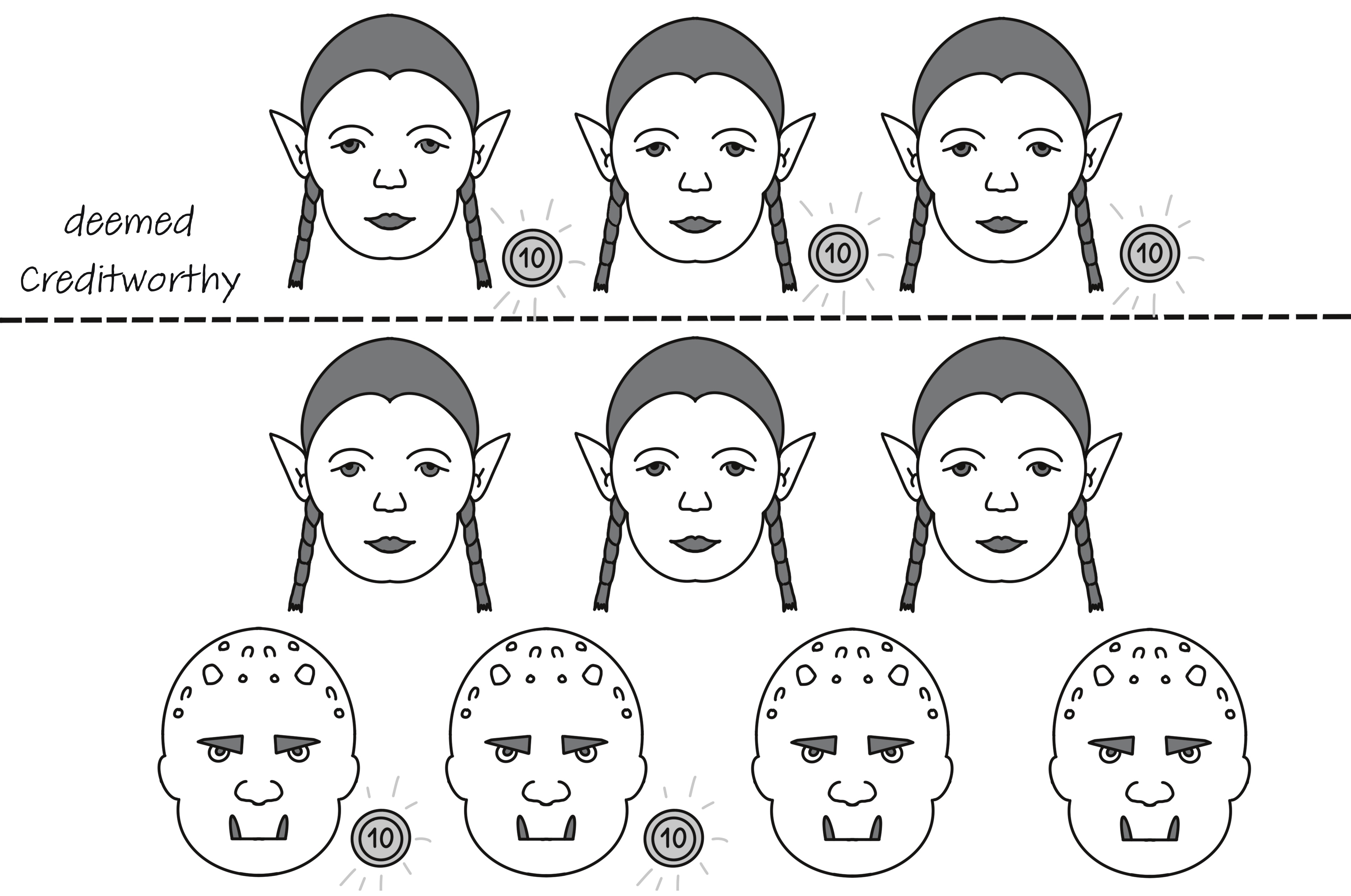}
    \label{dataset}
    \caption{A sample data set of creatures where only elves that earn more than 10 coins are considered creditworthy.}
\end{figure}

The bank director favors elves: Fig.~\ref{dataset} shows a typical credit decision by the bank. It can be clearly seen that ogres are discriminated against.

What is the goal of a governmental regulator in such a case? They want to equip the customers of the bank with an insight into the logic involved in the decision-making by the bank's ADM system, which is a black box. The regulator opts for a decision tree, learned by the CART algorithm with the Gini index. As a stopping criterion, they set a level $k$ such that the tree does not become too complex. They feel confident, as they have read that the relevance of an attribute in a learned decision tree can be quantified by two measures~\cite{freitas14}:
\begin{enumerate}
    \item The minimal depth in the tree at which the attribute first appeared;
    \item The total number of data points it was decisive for. 
\end{enumerate}
The regulator figures that all the important decisions are made in the first $k$ levels of the tree. They leave the decision of which input sample to learn the surrogate model from to the banks. 

A closer look reveals that this might be the entry door for manipulation: Three decision rules can be derived for the example in Fig.~\ref{dataset}: `If a creature has a salary of less than 10 coins per month, then they are to be classified as not creditworthy.', `If a creature has a salary of more than 10 coins a month and is an elf, then they are classified as creditworthy.' and `If a creature has a salary of more than 10 coins per month and is an ogre, then they are classified as not creditworthy.' The first two decision rules can be summarized into one rule: `If a creature has a salary of more than 10 coins per month and is an elf, then they are to be classified as creditworthy.'

\begin{figure}[h!]
    \centering
    \includegraphics[width=\textwidth]{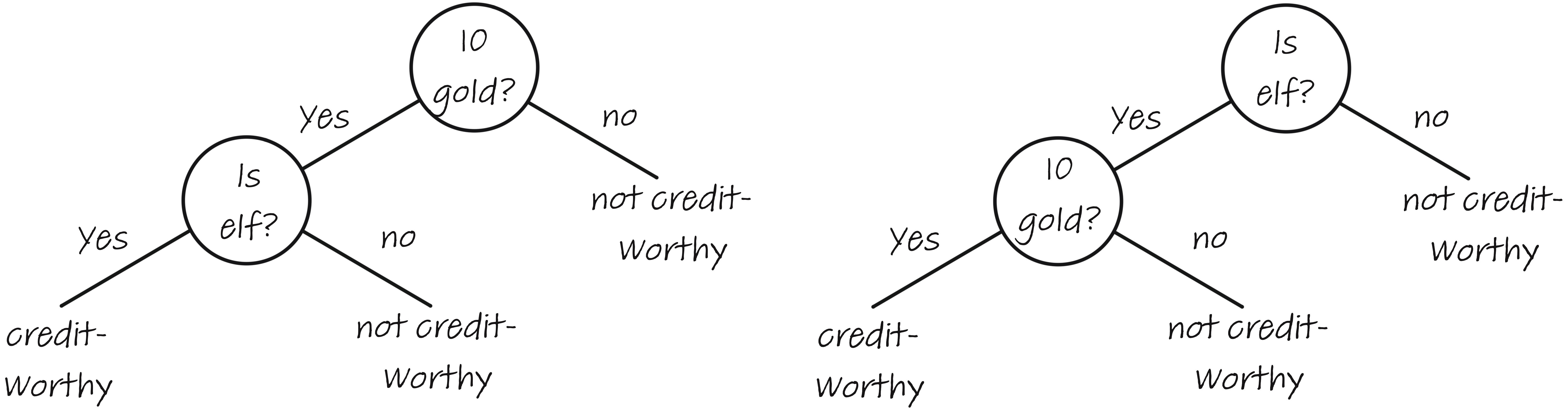}
    \label{treebspl}
    \caption{Two possible decision trees that show a decision about creditworthiness based on the species and salary of a creature.}
\end{figure}

However, Fig.~\ref{treebspl} shows that two perfect decision trees can be built from these three rules: One asks first for the species, one first for the salary.

The bank director asks themself whether they can take advantage of the freedom they have to create the input sample data set to their black box algorithm that will turn into the training data set of the surrogate model. They aim to find a data set that will push the sensitive attribute `species' to the level $k+1$ in the learned decision tree. They set out to understand which of the two trees would be learned from the data set in Fig.~\ref{dataset}.

A data point in this data set represents a creature which has two attributes: their salary, which will be classified as under or over 10 coins per month for simplification and their species, which is either elf or ogre. For this example, we consider a data set of ten creatures, six of whom are elves. Three of these six elves also have a salary over 10 coins. On the other hand, of the four ogres, two have a salary over 10 coins.

In this scenario, the bank classifies creatures as creditworthy if they have a salary of more than 10 coins and are an elf. Our data set therefore contains three creatures who are classified as creditworthy and seven creatures who are classified as not creditworthy, as seen in Figure~\ref{dataset}. Now, this is the basis of the surrogate model. For this example, the CART algorithm will stop when either level $k$ is reached or  all tuples in the remaining data set belong into the same class. Based on this, we can build the decision tree following algorithm~\ref{alg}.

At first the Gini index of the data set is calculated:
\begin{equation}\begin{split}
    \text{Gini }(y,S)=1-\bigg(\frac{3}{10}\bigg)^2-\bigg(\frac{7}{10}\bigg)^2=0.42
\end{split}\end{equation}
Since the Gini index is not $0$, we have not found a leaf node and continue.

Also, since the Gini index can have a maximum value of $0.5$, the level of heterogeneity in this example is relatively high. In particular, this means that if this data set as a whole were to be labeled as `creditworthy' or `not creditworthy', a high error rate is expected. If all data points were rated `not creditworthy', 30\% of them would be rated incorrectly.

In a next step, a splitting criterion is chosen from all possible attributes by calculating the Gini impurity of all of them. In this case, there are only two attributes and, therefore, two possible splits. First, the Gini impurity of `salary' is calculated:
\begin{equation}\begin{split}
    \text{G}&(\text{salary},S) = \sum_{i\in \text{dom}(\text{salary})}p_{i}\cdot \text{Gini }(y,S')\\
    &= 0.5\cdot \bigg(1-\bigg(\frac{3}{5}\bigg)^2-\bigg(\frac{2}{5}\bigg)^2\bigg)+0.5\cdot \bigg(1-\bigg(\frac{0}{5}\bigg)^2-\bigg(\frac{5}{5}\bigg)^2\bigg)\\
    &=0.5\cdot0.48+0.5\cdot0=0.24
\end{split}\end{equation}
Similarly, the Gini impurity for `species' is calculated as:
\begin{equation}\begin{split}
    \text{G}&(\text{species},S) = \sum_{i\in \text{dom}(\text{species})}p_{i}\cdot \text{Gini }(y,S')\\
    &= 0.6\cdot \bigg(1-\bigg(\frac{3}{6}\bigg)^2-\bigg(\frac{3}{6}\bigg)^2\bigg)+0.4\cdot \bigg(1-\bigg(\frac{0}{4}\bigg)^2-\bigg(\frac{4}{4}\bigg)^2\bigg)\\
    &=0.6\cdot0.5+0.4\cdot0=0.3
\end{split}\end{equation}
Because the Gini impurity of `salary' is lower, in the next step of the algorithm, we split the data set according to all values of `salary'. This results in two smaller data sets. In one there are only creatures who earn a salary under 10 coins per month. From these individuals, we already know that all of them are classified as not creditworthy. The Gini index for this data set is therefore $0$. Therefore, a leaf node is created containing the classification `not creditworthy'.

The other sub-data set $S'$ only contains creatures with a monthly salary of more than 10 coins. Among these, only the elves are considered creditworthy. Now  the Gini index for $S'$ is calculated again:
\begin{equation}\begin{split}
    \text{Gini }(y,S')=1-\bigg(\frac{3}{5}\bigg)^2-\bigg(\frac{2}{5}\bigg)^2=0.48
\end{split}\end{equation}
Here, we see a high degree of heterogeneity, which is to be expected.

Since we already used the attribute `salary' for a split, this data set can only be split further using the attribute `species' if it is required that each attribute is only used once. Therefore, the Gini impurity for the attribute `species' can now be calculated again.
\begin{equation}\begin{split}
    \text{G}&(\text{species},S') = \sum_{i\in \text{dom}(\text{species})}p_{i}\cdot \text{Gini }(y,S'')\\
    &= \bigg(0.6\cdot \bigg(1-\bigg(\frac{3}{3}\bigg)^2-\bigg(\frac{0}{3}\bigg)^2\bigg)+\bigg(0.4\cdot \bigg(1-\bigg(\frac{2}{2}\bigg)^2-\bigg(\frac{0}{2}\bigg)^2\bigg)\bigg)=0
\end{split}\end{equation}
The Gini impurity is $0$ and, therefore, has its minimal value. From this, we can conclude that splitting the data set based on the attribute `salary', two pure data sets will be formed.
A classification of these two subsets will lead to no errors. Therefore, two leaf nodes are reached in the next iteration of the algorithm.\\

This example shows how a decision tree can be built from a training data set. It produces the decision tree on the left in Fig.~\ref{treebspl}. Note, that the decision tree on the right in Fig.~\ref{treebspl} shows exactly the same classification, i.e., both representations are equivalent with respect to the classification. Which of them is built and then seen as \textbf{the} representation of the underlying decision-making process, however, is determined by the sample data set used as a training set for the CART algorithm. \textbf{In other words, if a different sample data set is used, a different decision tree is built.} Psychologically, it can be assumed that the right decision tree would be seen as more discriminating than the left one, while mathematically inclined readers will likely see that both decision trees will create discriminating results. Thus, if a malicious operator can influence what sample data set is used, they could manipulate the data set in a way that would hide discriminating decision trees in deeper levels of a decision tree, thus possibly deceiving normal customers.\\

In the next section, three theorems will be introduced, which show this idea.
\section{\label{order}Let There Be Order: How to Hack the Order of Attributes in a Decision Tree}
In the last section, an example showed that, for a data set with 60\% elves and 50\% of creatures with a salary over 10 coins, the attribute `salary' is first chosen in a decision tree. In this example, a creature has to be both an elf and have a high salary to be considered creditworthy. Both these values have, therefore, an influence in the decision on creditworthiness. However, in the decision tree, the attribute with the smaller percentage of the value that affects the creditworthiness in a favorable way, in this example the `salary', was chosen first. In this sense, let the appearance of an attribute be defined as the lowest level, at which it first appears in the tree (from root to leaves) and let the attributes be ordered by this.  

In this section, we will now show three theorems which help to generalize how the knowledge of the percentage of certain attributes in a sample population determines this order and, therefore, make it possible to hack decision trees using the sample data sets with which they are built. We start with a scenario in which the operator can choose the sample data set.

\subsection{Size Matters}
As in the last Section, for the following scenarios the operator can set their ADM system as they see fit. In this scenario, nobody of a disadvantaged subgroup is deemed creditworthy by the operator and all others get a credit if the values of all of their attributes come from a distinguished subset of favorable values. I.e., let $A_i^p\subseteq A_i$ be a distinguished set of values for the $i$-th attribute, these values are called the \textit{favorable values}. In our scenario here,  the operator will deem the $j$-th candidate from the data set as creditworthy iff $\forall i \in \left[1, \dots, n\right], x_j[i] \in A_i^p$, i.e., in each attribute they need to show a favorable value. The percentage of data points with a favorable value at attribute $i$ is denoted as $p_i$, i.e., $p_i := \frac{|\left\{x_j \mid x_j[i] \in A_i^p\right\}|}{n}$.

With this, we now prove the following theorem.
\begin{theorem}\label{theo1}
If the attributes are statistically independent of each other, $ p_{i'}  < p_{i''} < p_{i'''} <...$ determines the order in which the attributes are used in a decision tree from root to leaf. If they are dependent, for each subset of data points reaching the node, the attribute with $\min p_i$ in that set is chosen.
\end{theorem} 
First, this theorem is proven for two attributes and later generalized for any number of attributes.

We will now assume that the black box decision-making by the operator is audited by building a decision tree on the basis of some sample of data points with their attributes as input for the black box model and their label as output of the black box model. Let the percentage of people with a salary of ten coins be $p_s$. Similarly, let the percentage of elves be $p_e$. Both being an elf and having a salary of ten coins will be considered as favorable. Let now the surrogate model, i. e., the decision tree, be built on top of this data set.

To determine which attribute will be used in the root of a decision tree, Gini impurity of both attributes have to be compared. This comparison will be facilitated by simplifying the Gini impurity of each attribute first.

We can simplify the Gini impurity of the attribute `species' with a percentage of $p_e$ for its favorable value as:
\begin{equation}\begin{split}\label{equ:simplifiedspecies}
    \text{G}(species,S) &=\sum_{i\in \text{dom}(a)}p_{i}\cdot \text{Gini }(y,S')\\
    &=p_e\cdot(1-p_s^2-(1-p_s)^2)=2p_ep_s-2p_ep_s^2
\end{split}\end{equation}
Similarly, the Gini impurity of the attribute `salary' with an percentage $p_s$ for its favorable value can be simplified as:
\begin{equation}\begin{split}\label{equ:simplifiedsalary}
    \text{G }(salary,S) &=\sum_{i\in \text{dom}(a)}p_{i}\cdot \text{Gini }(y,S')\\
    &=p_s\cdot(1-p_e^2-(1-p_e)^2)=2p_ep_s-2p_e^2p_s
\end{split}\end{equation}
For one attribute to be chosen in the root of the decision tree, its Gini impurity has to be lower than the Gini impurity of the other attribute. The difference of the Gini impurities can be used to compare them, as seen in equation~\ref{equ:differencetwoattributes}.
\begin{equation}\begin{split}\label{equ:differencetwoattributes}
    \triangle(\text{G}(species,S)-\text{G}(salary,S))&=(2p_ep_s-2p_ep_s^2)-(2p_ep_s-2p_e^2p_s)\\
    &=2p_e^2p_s-2p_ep_s^2=2p_ep_s\cdot(p_e-p_s)
\end{split}\end{equation}
If this difference is negative, the Gini impurity of `species' is lower, and this attribute is therefore chosen first when building a decision tree. Alternatively, the attribute `salary' is chosen first if the difference is positive.

For the difference to be negative, the percentage of elves $p_e$ has to be lower than the percentage of creatures who have a high salary and vice versa. Therefore, the attribute `salary' is chosen in the root of a decision tree, if $ p_e > p_s$. The attribute `species' is then chosen in depth $k=2$ of the decision tree. This proves theorem~\ref{theo1} for two attributes.

\begin{figure}[h!]
    \centering
    \includegraphics[width=0.9\textwidth]{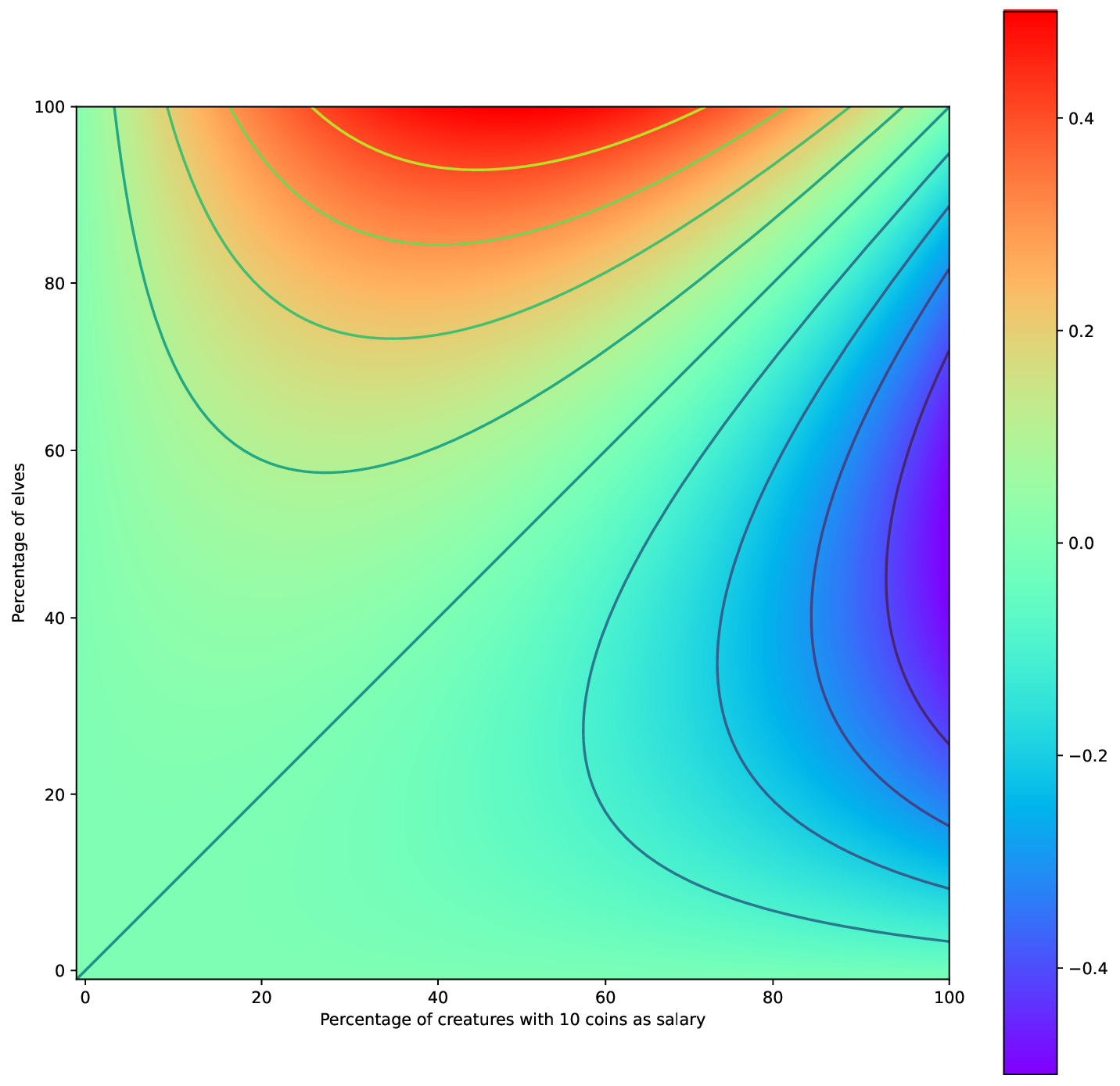}
    \label{difference}
    \caption{Difference of the Gini impurities of the attribute ‘species’ and ‘salary’. Where the difference is positive, the attribute ‘salary’ would be chosen in the root of a decision tree.  Where  the  difference  is  negative,  the  attribute  ‘species’  would  be  chosen, respectively.}
\end{figure}

Fig.~\ref{difference} shows how different percentages of elves and creatures with a salary of ten coins affect the difference of both Gini impurities. The area above the diagonal shows the area where the difference is positive and, therefore, the attribute `salary' is chosen as the root of a decision tree. The area below the diagonal shows, respectively, where the attribute `species' would be chosen first.
\hfill\\
After showing that this theorem holds for two attributes, we now generalize it for more than two attributes.

The percentage of data points with a favorable value of attribute $j$ will be shortened as $p_j$ for simplification. Then, we can simplify its Gini impurity as seen in eq.~\ref{equ:simplifiedmultiattributes}.
\begin{equation}\begin{split}\label{equ:simplifiedmultiattributes}
    \text{G}(a_k,S) &=\sum_{j\in \text{dom}(a)}p_{j}\cdot \text{Gini }(y,S')\\
    &=2\cdot\prod_{j\in I}p_j-2\cdot p_k\cdot\prod_{j\in I\backslash\{k\}}(p_j)^2 
\end{split}\end{equation}
\textbf{Derivation:}\\
\begin{equation*}\begin{split}
    \text{G}(a_k,S) &=\sum_{j\in \text{dom}(a)}p_{j}\cdot \text{Gini }(y,S')\\
    &=p_k\cdot(1-(\prod_{j\in I\backslash\{k\}}p_j)^2-(1-\prod_{j\in I\backslash\{k\}}p_j)^2)\\&+(1-p_k)\cdot(1-(0)^2-(1-1)^2)\\
    &=p_k\cdot(1-(\prod_{j\in I\backslash\{k\}}p_j)^2-(1-2\cdot\prod_{j\in I\backslash\{k\}}p_j+(\prod_{j\in I\backslash\{k\}}p_j)^2))\\
    &=p_k\cdot(2\cdot\prod_{j\in I\backslash\{k\}}p_j-2\cdot(\prod_{j\in I\backslash\{k\}}p_j)^2)\\
    &=2\cdot\prod_{j\in I}p_j-2\cdot p_k\cdot\prod_{j\in I\backslash\{k\}}(p_j)^2
\end{split}\end{equation*}
Similarly to the previous case with two attributes, we can find the attribute which will be used in the root of the decision tree by searching for the attribute with the lowest Gini impurity. The Gini impurity of two arbitrary attributes $o$ and $q$ and a data set with percentages of $p_o$ and $p_q$ for their favorable value, respectively, can be compared by using their difference, as seen in eq.~\ref{equ:differencemultipleattributes}.
\begin{equation}\begin{split}\label{equ:differencemultipleattributes}
     \triangle(\text{G}(o,S)&-\text{G}(q,S))\\ &=2\cdot p_o\cdot p_q\cdot\prod_{i\in I\backslash\{o,q\}}(p_i)^2 \cdot (p_o-p_q)
\end{split}\end{equation}
\textbf{Derivation:}\\
\begin{equation*}\begin{split}
\triangle(\text{G}(o,S)&-\text{G}(q,S))=(2\cdot\prod_{i\in I}p_i+2\cdot p_o\cdot\prod_{i\in I\backslash\{o\}}(p_i)^2)\\
&-(2\cdot\prod_{i\in I}p_i+2\cdot p_q\cdot\prod_{i\in I\backslash\{q\}}(p_i)^2)\\
&=2\cdot p_o\cdot\prod_{i\in I\backslash\{q\}}(p_i)^2-2\cdot p_q\cdot\prod_{i\in I\backslash\{o\}}(p_i)^2\\
&=2\cdot p_o\cdot p_q\cdot\prod_{i\in I\backslash\{o,q\}}(p_i)^2 \cdot (p_o-p_q)\\
\end{split}\end{equation*}
If the difference is positive, the attribute $q$ is chosen in the root of the decision tree. Alternatively, if the difference is negative, the attribute $o$ is chosen in the root. Again, whether an attribute is chosen first is determined by the size of $p_o$ and $p_q$. This is similar to the previous case with only two attributes. The smallest percentage of a favorable value determines that this attribute is chosen first. This holds for all pairs of attributes $a_1,..., a_n$ where $a_i\neq a_j$.

Let $a_1$ be chosen as the attribute in the root of the decision tree w.l.o.g. Then, all data points in the second level of the tree are either not creditworthy and have a value that is not favorable, or have to be split further into groups to determine their creditworthiness and have the favorable value for the first chosen attribute. For the later data points, we can determine the Gini impurity of all remaining attributes with equation~\ref{equ:simplifiedmultiattributes} and~\ref{equ:differencemultipleattributes}. We can easily see that neither the simplification for the Gini impurity nor the difference of Gini impurity do structurally change. If we would simplify them for each level of the decision tree, we would arrive at eq.~\ref{equ:simplifiedspecies},~\ref{equ:simplifiedsalary}  and~\ref{equ:differencetwoattributes} for the Gini impurities of the last two attributes.

Therefore, if all attributes are independent from each other and $p_1 < p_2 < … < p_{n-1} < p_n$, the attribute $a_1$ is chosen in the root node, the attribute $a_2$ in depth 2 and so forth. For dependent attributes, the smallest $p_i$ could change in each node, dependent on the attribute chosen in the previous node. Therefore, $\min p_i$ is chosen in each node of the decision tree. This proves theorem~\ref{theo1}.\\

Using the theorem proven in this section, we can derive the depth of an attribute in a decision tree by its percentage of favorable values in the data set without calculating the Gini impurity for each attribute.

Once the regulators notice this patterns, they are not amused. From now on, they will determine the sample data set that is fed to a black box ADM system. They can, e.g., determine that the percentage of ogres and elves need to be similar to that in the full population or even require them to be balanced in the sample data set. In the next subsection, we will prove a theorem that gives the percentage of the favored group that can still be favored with credits under any given percentage of ogres. 

\subsection{Let There Be Ogres}

Again, a malicious operator wants to discriminate against a certain group using an ADM system. The decision rules of this ADM system follow the same principle as before, meaning that a person is only considered creditworthy iff they possess only favorable values for all their attributes. In the following scenario, however, the operator cannot choose the data set to train a legally required decision tree. Instead, a data set is defined or even provided by regulators, which is meant to be labeled by the ADM system; this is then used to train a decision tree.

In this scenario, a malicious operator would have to manipulate their ADM system to hide discrimination in a training decision tree. Again, the operator wants none of the disadvantaged group to be classified as creditworthy. Based on the percentage of favorable values in the provided data set, we will derive the percentage of the advantaged group the operator can favor while the decision tree will not ask for the sensitive attribute until the last level of the tree.
\begin{theorem}\label{theo2}
If there are $x\%$ of the disadvantaged group and the advantaged group is strictly bigger than any other group of favorable value and the attributes are statistically independent, $(100-x)^k\%$ of data points can still be favored in a decision and the decision tree will not ask for the sensitive attribute until the last level of the decision tree.
\end{theorem}
In the last section and for statistically independent attributes, we have shown that $ p_{i'} < p_{i''} < p_{i'''} <...$ gives the order of attributes in a decision tree, with $p_{i'}$ being chosen in the root of a decision tree. Because the attributes are statistically independent, the percentage of $p_{i}$ is the same in each level of the tree, meaning that in level $k$ the percentage of an arbitrary value $v_{i}$ of an attribute is the same as in the whole data set if it has not yet been chosen in an earlier level of the decision tree.

Additionally, the decision in the first level of the tree splits the data set into two smaller data sets. $(1-p_{i'})$ percent of the data set will be classified as `not creditworthy', and the remaining $p_{i'}$ percent are a mixed group of `creditworthy' and `not creditworthy' data points. The later data set will then be split further by the next decision question until the branch that asks for the favorable values of all attributes only consists of data points that are considered `creditworthy'. Therefore, in level $2$, only $p_{i'}$ percent of the original data set will be split further. Of these $p_{i'}$ percent, $p_{i''}$ percent will split further in layer $3$ and so forth. Based on this, the exact percentage of positive classified data points is $ p_{i'} \cdot p_{i''} \cdot p_{i'''} \cdot …\cdot p_{i^n}$, with $p_{i^n}$ being the percentage of data points with a favorable value of the attribute that is used as the last split in the decision tree.

In this scenario, a malicious actor would want the discriminating decision rule as deep in the decision tree as possible. Therefore, $p_{i^n}$ would be the percentage of data point that have a favorable value for the sensitive attribute of species. According to Theorem~\ref{theo1}, all other percentages of data points with favorable values for other attributes have to be smaller.

So, in each level of the decision tree, a percentage $p_{i^l}<p_{i^n}$ of data points with a favorable value for the decision question in this level will be given to the next level until the last level is reached. Then, $(p_{i^n})^k$ is the upper bound of data points that can be classified as creditworthy if there are $k$ attributes. And by implication, if the percentage of the disadvantaged group is given by $x=(1-p_{i^n})$, the upper bound is $(1-x)^k$ in relation to the percentage data points that belong to the disadvantaged group.

This proves Theorem~\ref{theo2}.\\

\begin{figure}[h!]
    \centering
    \includegraphics[width=0.8\textwidth]{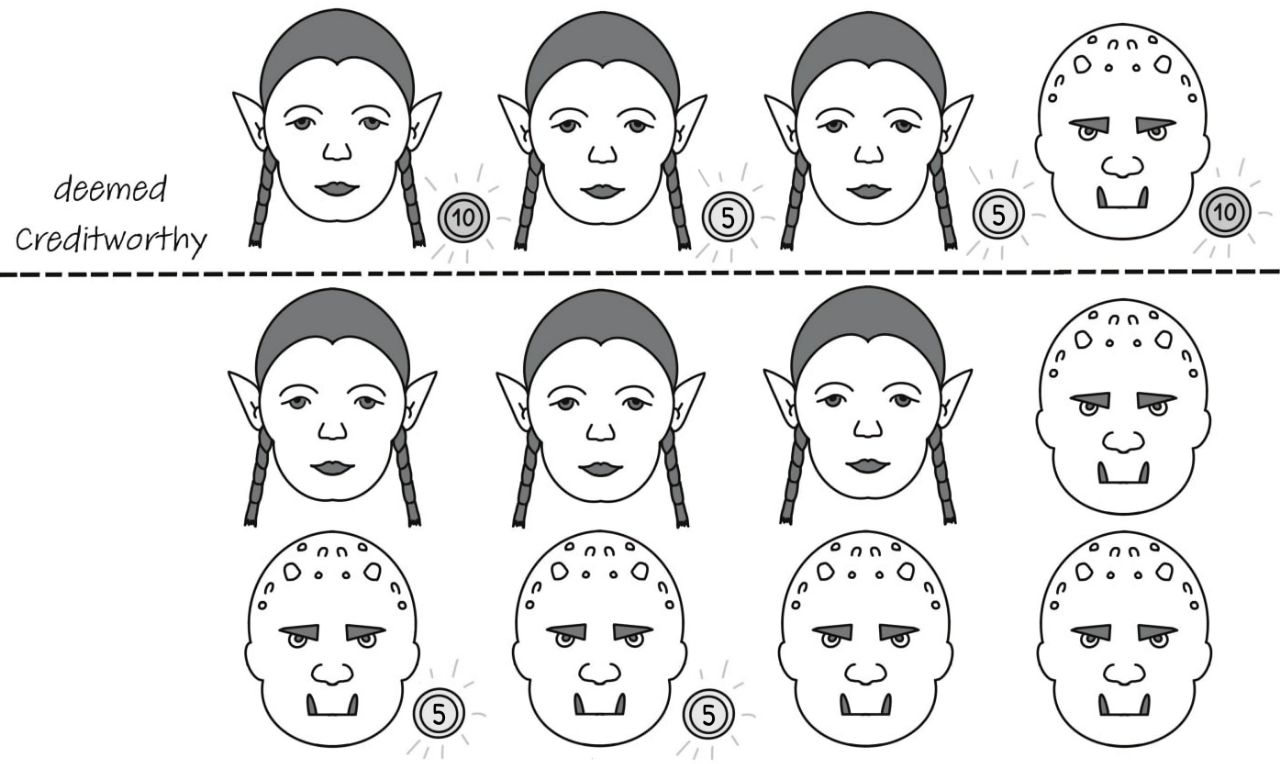}
    \label{dataset2}
    \caption{ A data set of creatures where all creatures with a salary of 10 coins and elves with salary of 5 coins are considered creditworthy.}
\end{figure}

Both Theorem~\ref{theo1} and~\ref{theo2} can help a malicious operator to hide discrimination of a certain subgroup in lower levels of a decision tree. While this is effective when only taking the decision tree into account, decision trees built in these scenarios can easily be sanity checked by calculating the total percentage of a subgroup that is deemed creditworthy in these decision trees or the ADM system, given a random data set. Since no one in the disadvantaged subgroup is deemed creditworthy, such discrimination cannot be feasibly hidden.
Therefore, for the next scenario, the point of view of the regulator, who doubts if decision trees are a useful tool to showcase and search for such discrimination, is considered. 

\subsection{Casting Doubt}
In the last scenario, the regulator has seen that there are cases in which discrimination is not clearly visible. Because of this, they ask themself how pronounced discrimination has to be for the sensitive attribute to be in the first level of the decision tree.

In this scenario, the regulator assumes an optimal sample data set in which the groups of ogres and elves are balanced. They also assume that, realistically, a bank would consider any creature with a significantly high enough salary as creditworthy, since the risk that such a creature would not be able to pay back their loan is considerably low. Additionally, a malicious operator wants to discriminate against ogres. Because of this, elves need a lower salary compared to ogres to be considered creditworthy.

Therefore, the attribute `salary' is split into three groups. With a low salary of zero coins, no creatures are considered creditworthy, with a medium salary of five coins, only elves are considered creditworthy, and with a high salary of ten coins, every creature is considered creditworthy. 

Then, the question of the regulator is how big the percentage of creatures with a medium salary can be for the decision tree to not ask for the sensitive attribute in the first level of the tree.

Figure~\ref{dataset2} shows a possible data set in this scenario and its labels from such an ADM system.

\begin{theorem}\label{theo3}
If the disadvantaged and the advantaged group are of the same size and the attribute values are independent of each other, up to $50$\% of all data points can have a medium salary, and the decision tree will not ask for the sensitive attribute in the first level of the tree.
\end{theorem}

Let $p_0$, $p_5$ and $p_{10}$ be the percentage of creatures with a low, medium and high salary, respectively. With $p_e=p_5=0.5$, the percentage of elves, and $p_{10}=1-p_0-p_5$ the Gini impurity for `species' can be simplified to:
\begin{equation}\begin{split}
    \text{G}(species,S) &=\sum_{i\in \text{dom}(a)}p_{i}\cdot \text{Gini }(y,S')\\
    &=p_e\cdot(1-p_0^2-(1-p_0)^2)+(1-p_e)\cdot(1-p_{10}^2-(1-p_{10})^2)\\
    &=2p_0p_e\cdot (1-p_0)+2p_{10}\cdot(1-p_e)\cdot(1-p_{10})\\
    &=0.25+p_0\cdot (1-2p_0)
\end{split}\end{equation}
For the attribute ‘salary’, three possible splits can be performed in a binary decision tree\footnote{For a non binary decision tree, the Gini impurity for ‘salary’ is $\text{G}(salary,S)=p_0\cdot (0)+ p_5\cdot (1- p_e^2-(1-p_e)^2)+p_{10}\cdot (0)=2p_ep_5\cdot (1-p_e)=0.25$. The theorem also holds in this case.}, according to their salary. For `low salary', i.e., the decision question splits the data set into a subset of data points with low salary and a subset of data points with middle and high salary.

These Gini impurities can be simplified to: 
\begin{equation}\begin{split}
    \text{G}(low \text{ }salary,S)  &=\sum_{i\in \text{dom}(a)}p_{i}\cdot \text{Gini }(y,S')\\
    &= (p_5+p_{10})\cdot\bigg(1-\bigg(\frac{p_5\cdot(1-p_e)}{p_5+p_{10}}\bigg)^2-\bigg(1-\frac{p_5\cdot(1-p_e)}{p_5+p_{10}}\bigg)^2\bigg)\\
    &+p_0\cdot\bigg(1-\bigg(\frac{p_0}{p_0}\bigg)^2-\bigg(1-\frac{p_0}{p_0}\bigg)^2\bigg)\\
    &=2p_5\cdot(1-p_e)\cdot\bigg(1-\frac{p_5\cdot(1-p_e)}{p_5+p_{10}}\bigg)\\
    &=0.5-\frac{0.125}{1-p_0}
\end{split}\end{equation}
\begin{equation}\begin{split}
    \text{G}(high\text{ }salary,S)  &=\sum_{i\in \text{dom}(a)}p_{i}\cdot \text{Gini }(y,S')\\
    &= (p_0+p_5)\cdot\bigg(1-\bigg(\frac{p_5\cdot (1-p_e)}{p_0+p_5}\bigg)^2-\bigg(1-\frac{p_5\cdot (1-p_e)}{p_0+p_5}\bigg)^2\bigg)\\
    &+p_{10}\cdot\bigg(1-\bigg(\frac{p_{10}}{p_{10}}\bigg)^2-\bigg(1-\frac{p_{10}}{p_{10}}\bigg)^2\bigg)\\
    &=2p_5\cdot(1-p_e)\cdot\bigg(1-\frac{p_5\cdot(1-p_e)}{p_0+p_5}\bigg)\\
    &=0.5-\frac{0.125}{p_0+0.5}
\end{split}\end{equation}
Note that $\text{G}(low \text{ }salary,S)$ and $\text{G}(high \text{ }salary,S)$ are equal, if $p_0=p_{10}$.
\begin{equation}\begin{split}
    \text{G}(medium \text{ }salary,S)  &=\sum_{i\in \text{dom}(a)}p_{i}\cdot \text{Gini }(y,S')\\
    &= p_5\cdot(1-p_e^2-(1-p_e)^2)\\
    &+(p_0+p_{10})\cdot\bigg(1-\bigg(\frac{p_{0}}{p_0+p_{10}}\bigg)^2-\bigg(1-\frac{p_{0}}{p_0+p_{10}}\bigg)^2\bigg)\\
    &=2p_5p_e\cdot(1-p_e)-2p_0\cdot(1-\frac{p_0}{p_0+p_{10}})\\
    &=0.25+2p_0\cdot(1-2p_0)
\end{split}\end{equation}
\begin{figure}[h!]
    \centering
    \includegraphics[width=\textwidth]{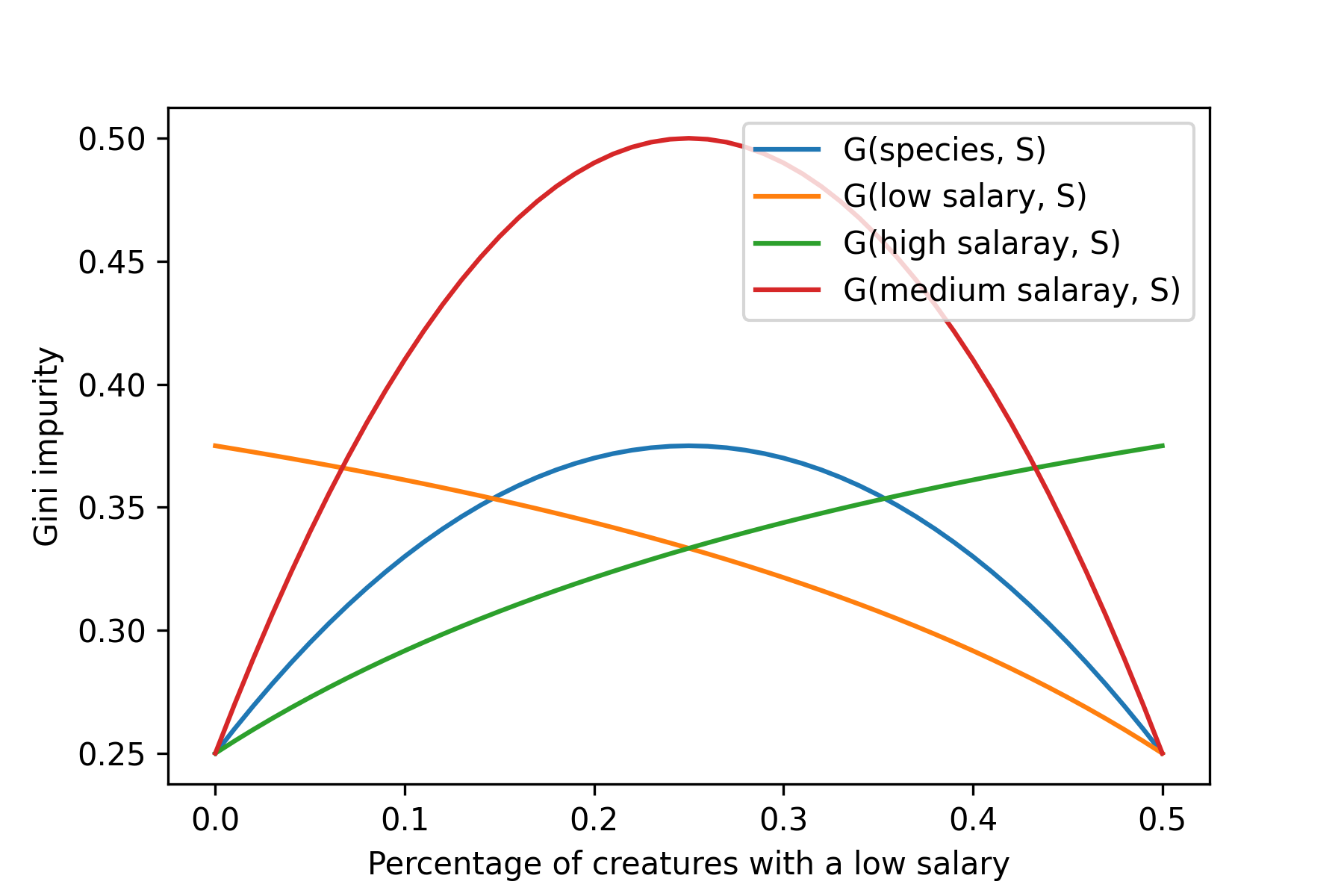}
    \label{middleincome}
    \caption{Height  of  Gini  impurity,  if  only  ogres  with  a  medium  salary  are  discriminated against. In this case, 50\% of the data set have a medium salary and 50\% are elves. The attribute with the smallest Gini impurity is chosen in the root node of a decision tree, meaning that the sensitive attribute species can only be chosen, if there are no creatures with high or low salary. In these two cases, all Gini impurities are equal.}
\end{figure}

In a next step, the Gini impurities are compared by finding the intersections of the Gini impurities in the interval from $0$ to $0.5$. Since it holds that $p_5=0.5$, which is $50$\% of the data set, values higher than $0.5$ for $p_0$ are not possible.

For $\text{G}(medium \text{ }salary,S)$ and $\text{G}(species,S)$, it is easily seen that both are only equal for $p_0=0$ and $p_0=0.5$. Except for these corner cases, where there are no creatures with low or high income, $\text{G}(middle \text{ }salary,S)$ is higher and, therefore, the attribute `species' would be preferred as a split for the root node of a decision tree.

$\text{G}(low \text{ }salary,S)$ and $\text{G}(species,S)$ have intersections in points $(0.146|0.354)$ and $(0.5|0.25)$. Between these two points, `low salary' has a smaller Gini impurity. For $p_0 < 0.146$, $\text{G}(low \text{ }salary,S)$ has a smaller value.

Similarly, $\text{G}(high \text{ }salary,S)$ and $\text{G}(species,S)$ have intersections in points $(0|0.25)$ and $(0.354|0.354)$. Between these two points `high salary' has a smaller Gini impurity.

Since in the interval from $0$ and $0,354$ 'high salary' has a smaller Gini impurity than `species', and in the interval from $0,146$ to $0,5$ `low salary' has a smaller Gini impurity than `species', the attribute `species’ will not be chosen in the root node of a decision tree, see Figure~\ref{middleincome}. The only exception of this are the corner cases, for which all Gini impurities are the same and, therefore, all attributes could be chosen for a split. Note that these corner cases are similar to the scenario discussed in Theorem~\ref{theo1}.

\begin{figure}[h!]
    \centering
    \includegraphics[width=\textwidth]{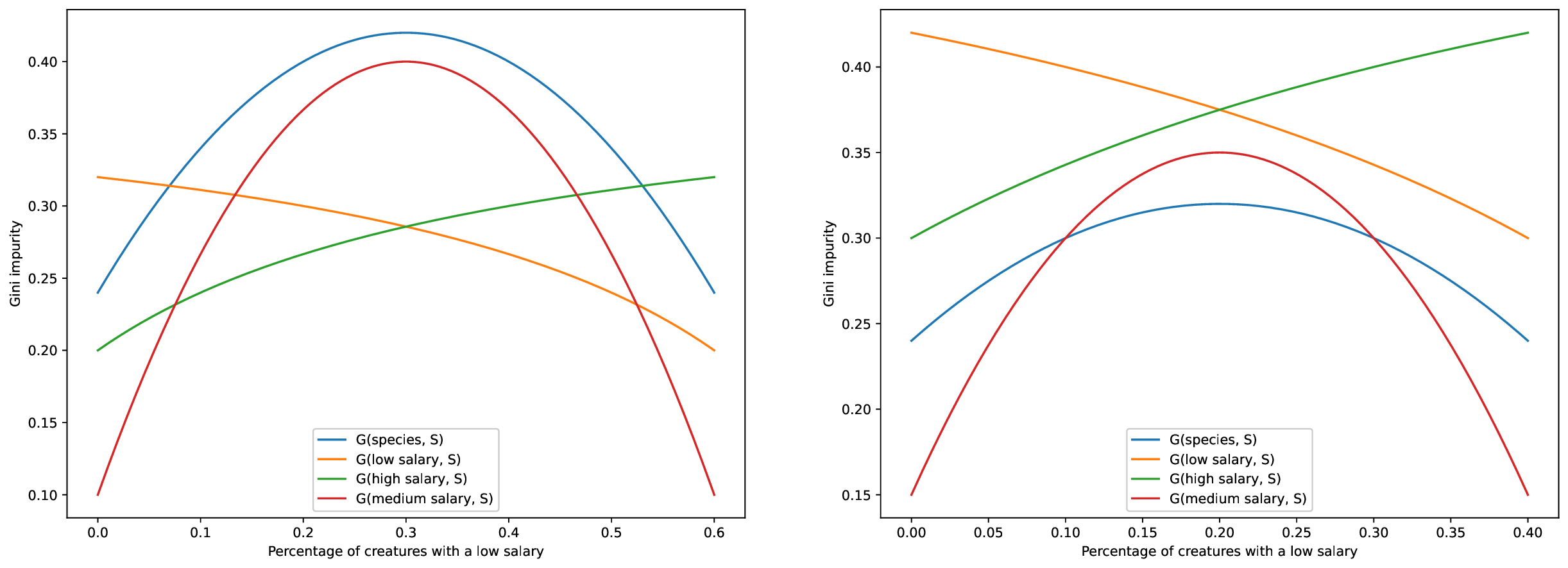}
    \label{varmiddle}
    \caption{Height  of  Gini  impurity,  if  only  ogres  with  a  medium  salary  are  discriminated against. In the first case, 40\% of the data set have a medium salary, in the second case 60\% of the data set have a medium salary, and in both cases 50\% are elves. The lowest Gini impurity will be chosen in the root node of a decision tree.}
\end{figure}

For a smaller percentage of creatures with medium salary, $\text{G}(species,S)$ is greater than at least one Gini impurity of salary in all cases. For a higher percentage, it is smaller than all other Gini impurities, see Fig.~\ref{varmiddle}. The attribute `species' gains, therefore, importance if the percentage of data points where creatures are classified different according to their species is higher. If this percentage is maximal, the attribute `species' would be the only factor influencing the creditworthiness. If this percentage is minimal, no discrimination occurs. This also implies that if the percentage of the disadvantaged group which is classified negatively based on their species is small, this discrimination is likely to go unnoticed.

Theorem~\ref{theo3} shows that for scenarios where only a fraction of a disadvantaged group is discriminated against, this fraction has to be relatively high to show the sensitive value in the root of a decision tree. This is especially relevant, because it makes it harder to detect such discrimination, especially since small disadvantages for certain subgroups might be inevitable, e.g. because of poorly chosen training data sets for the ADM system, the fact that ogres did not apply for credits often in the past, and so on. In these cases, the discriminating decision rules might be hidden in deeper levels of a decision tree and also harder to detect in a data set, since the percentage of disadvantaged creatures is smaller.

The scenario used for Theorem~\ref{theo3} is more realistic than the examples seen before, since even a malicious actor might consider ogres with a high salary as creditworthy solely because of economical reasons. But, such discrimination might also occur if the operator is not malicious but lawful and simply chooses a poor training data set for their ADM system, or argues that, because ogres have not paid back their loans in the past, the bank can only grant credits if they have an additional security in form of clients with a higher salary in this subgroup.

We also argue that such discrimination could be even more hidden if there are more attributes involved as in the scenario shown here. More attributes involved in the decision of creditworthiness could push the sensitive attribute further down in a decision tree, especially if these attributes influence the decision for all subgroups equally.

On the other hand, for a regulator who wants to use a decision tree as a surrogate model for an unknown black box model, this means that inequality that is not as strongly pronounced might go unnoticed if a decision tree is not carefully examined. More importantly, it has to be considered that a sensitive attribute might seem of low importance because its position in the decision tree, even if it has a relatively strong impact on the classification. This makes it harder for a regulator to trust the decision tree as a surrogate model, and even raise doubts whether the model is useful if it is perceived as ``inaccurate'' based on the feeling that a sensitive attribute should be of greater importance as its position in the decision tree suggests. In this case, a regulator might not feel confident to make assumptions about the involved logic of a black box model on the basis of only the surrogate model.\\

In the next section, implication of the three theorems will be further discussed.

\section{\label{Implications}(Real life) Implications}
In the last section, we have seen that, by altering the properties of the training data set of a decision tree, we can change the order of appearance of attributes in the decision tree. Different properties result in a different order of decision rules inside the decision tree, and it is possible to find regularities in the order of attributes in this type of decision trees. Using these regularities, malicious actors could hide discriminating decision rules in lower levels of the decision tree by altering the data set used to train the decision tree or influencing the black box with which the training data set is labeled. However, benevolent actors, e.g. quality assurance or regulators, could also make them more visible.

The examples and rule set of the black box model to determine who is considered creditworthy used in this paper are simplified. We argue that determining creditworthiness, even in a fictitious society, is a much more complex task. Discrimination against ogres as a group of creatures might be less blatant, e.g. as seen in the scenario of Theorem~\ref{theo3}, and creditworthiness might be affected by other features making its calculation more difficult. I.e. a bank might consider someone as creditworthy if this person is pursuing a higher education or has a house, even if their salary is relatively low. The size and repayment method of a loan might also be put into consideration.

Additionally, in this example, creatures were only discriminated against based on their species. Other factors might also be a cause of discrimination, e.g. their social class, disabilities, age, gender, or other factors. A black box model could also discriminate against someone based on their social status and against another person based on their gender. Because both of these forms of discrimination exist inside the model, it might be harder to identify them or evaluate their impact inside a decision tree or any other explainability approach.\\

While more complex rules make a decision tree potentially harder to read, the measurement of the importance of an attribute might help to grasp the relevance of certain rules inside the classification process.

The relevance or importance of an attribute in a classification process is usually determined by two factors: the depth of the appearance of this attribute and the number of instances in which this attribute influences the classification of data points~\cite{freitas14}. 

For the examples shown in this paper, we know that all attributes contribute evenly in the classification process, meaning all of them are equally important. Every attribute has to be used inside a decision tree to achieve a classification free of errors for every data point. Furthermore, all of them are potentially equally important in this classification process. This can be easily seen by giving all favorable values an equal percentage. In this case, all Gini indices are the same for all attributes, meaning the decision which attribute is chosen in a given depth is dependent on the implementation of the decision tree algorithm. Similarly, in each level of the tree, a percentage of data points is classified, meaning decision questions in deeper levels of the tree affect less data points. Because of this, both measures of relevance for these attributes fail. The structure of a decision tree does not allow for attributes to be of equal importance. Since there can only be one root node and this root node splits the data set into smaller fractions in its branches, one attribute is of usually greater importance than all other attributes. Furthermore, both measures give attributes closer to the root potentially greater and attributes with higher depth potentially lower relevance. Since we have seen that the depth of attributes can be manipulated by changing the properties of the training data set of the decision tree, this relevance can be directly influenced by a malicious actor. Attributes which clearly point towards discriminating rules inside the black box can be pushed further down inside the decision tree giving them a perceived lower importance or relevance inside the model. Alternatively, in other examples they could be given a perceived greater importance to sabotage the reputation of a given black box model.

While discrimination cannot be completely hidden by this approach, a low relevance of such decision rules might be deemed acceptable by societal standards because they might result from outside factors, i.e., discrimination against ogres might be the result of ogres not being able to pay back their loans in the past. Furthermore, if a low relevance of discriminating decision rules is deemed acceptable, models which are represented by decision trees, like seen in this paper, might be accepted by society even if no ogres will ever get a loan.

\subsection*{\label{futureresearch}Further Research}
Most scenarios and examples used in this paper are minimal in their attribute number and the underlying ADM system. Further research is needed to determine if similar regularities to exploit a decision tree as an approach for explainable AI can be found if the underlying ADM system is more complex in its decision rules or more attributes are involved. This research is especially crucial,
because in these scenarios, a decision tree is of much higher depth and, therefore, more complex. An operator or regulator might be tempted to limit the size of the decision tree in these cases to improve readability. Then, discriminating decision rules might be omitted because of the stopping criteria used. A better understanding of the relationship between a black box, the labeled training data set of a decision tree, and a decision tree helps to determine when a trade-off between readability and precision in decision trees should be considered, and when not.

Similarly, we assumed that attribute values are independent of each other. In reality, this is often not the case. Minorities might also have to deal with lower salaries and, therefore, the salary distribution for the advantaged and disadvantaged groups are not equal. Further research has to determine if and how a different distribution of certain attributes could affect the order of attributes in decision trees.

\section{\label{conclusion}Conclusion}
In this paper, we have shown that certain regularities exist which determine the order of appearance of attributes in a decision tree. There are scenarios in which these regularities can be used to hide an unfair discrimination or to favor a group unfairly without the risk of identification of these issues. We argue that regularities like these could be found and influence more complex and true to life decision trees. These regularities could be exploited to hide the existence of discriminating decision rules in deeper levels of decision trees or make them more apparent.

This paper outlines possible implications of those regularities on a fictitious society, but we argue that they could be easily transferred to real life examples. By replacing the attribute of species with, e.g., the attribute of gender we can easily transfer the examples shown in this paper to more realistic scenarios. Of course, this also means the outlined implications of manipulating a training data set to rearrange the importance of certain attributes in a decision tree are also valid for real life examples. This could mean that black box models which contain harmful biases against certain groups would be accepted and used in today's society and perpetuate those biases in future decision-making.

Conversely, regularities as shown in this paper could also improve our understanding of decision trees making them a more transparent and accepted approach for explainable AI.

\section*{Abbreviations}
ADM, Algorithmic Decision-Making; AI, Artifical Intelligence; ML, Machine Learning; CNF: Conjunctive Normal Form

\backmatter

\bmhead{Declarations}

\section*{Availability of data and materials}
Not applicable

\section*{Competing interests}
The authors declare that they have no competing interests.

\section*{Funding}
This research was funded by the Volkswagen Foundation within the project 
``Deciding about, by, and together with algorithmic decision-making systems'' (2019-2023).

\section*{Authors' contributions}
KAZ conceived the initial idea of this research and developed the research question together with AW; AW performed the research and wrote the article; and KAZ and AW edited the draft. All authors read and approved the final manuscript.

\section*{Acknowledgments}
Not applicable


\bibliography{main}

\end{document}